 \documentclass[pmlr,twocolumn,10pt,table]{jmlr} 

\usepackage{booktabs}
\usepackage{siunitx}

\usepackage{dsfont}

\usepackage[format=plain,labelfont=bf,labelsep=colon]{caption}
\usepackage{multirow}
\usepackage{enumitem} 

\usepackage[switch]{lineno}

\theorembodyfont{\upshape}
\theoremheaderfont{\scshape}
\theorempostheader{:}
\theoremsep{\newline}

\jmlrvolume{297}
\jmlryear{2025}
\jmlrworkshop{Machine Learning for Health (ML4H) 2025} 

 \title[Classifying Phonotrauma Severity with Soft Ordinal Regression]{Classifying Phonotrauma Severity from Vocal Fold Images with Soft Ordinal Regression}

\author{%
\Name{Katie Matton} \Email{kmatton@mit.edu}\\
\Name{Purvaja Balaji} \Email{pbalaji@mit.edu}\\
\addr Massachusetts Institute of Technology
\AND
\Name{Hamzeh Ghasemzadeh} \Email{hamzeh.ghasemzadeh@ucf.edu}\\
\addr School of Communication Sciences and Disorders, University of Central Florida
\AND
\Name{Jameson C. Cooper} \Email{jcooper12@mgh.harvard.edu}\\
\Name{Daryush D. Mehta} \Email{mehta.daryush@mgh.harvard.edu}\\
\Name{Jarrad H. Van Stan} \Email{jvanstan@mgh.harvard.edu}\\
\Name{Robert E. Hillman} \Email{hillman.robert@mgh.harvard.edu}\\
\addr Center for Laryngeal Surgery and Voice Rehabilitation, Massachusetts General Hospital \\
\addr Department of Surgery, Harvard Medical School \\
Massachusetts General Hospital Institute of Health Professions
\AND
\Name{Rosalind Picard} \Email{picard@mit.edu}\\
\Name{John Guttag} \Email{guttag@mit.edu}\\
\addr Massachusetts Institute of Technology
\AND
\Name{S. Mazdak Abulnaga} \Email{abulnaga@csail.mit.edu}\\
\addr Athinoula Martinos Center, Massachusetts General Hospital,
Harvard Medical School \\
\addr Massachusetts Institute of Technology
}

\begin{document}

\maketitle

\begin{abstract}
Phonotrauma refers to vocal fold tissue damage resulting from exposure to forces during voicing. 
It occurs on a continuum from mild to severe, and treatment options can vary based on severity. 
Assessment of severity involves a clinician’s expert judgment, which is costly and can vary widely in reliability. 
In this work, we present the first method for automatically classifying phonotrauma severity from vocal fold images.
To account for the ordinal nature of the labels, we adopt a widely used ordinal regression framework.
To account for label uncertainty, we propose a novel modification to ordinal regression loss functions that enables them to operate on soft labels reflecting annotator rating distributions.
Our proposed soft ordinal regression method achieves predictive performance approaching that of clinical experts, while producing well-calibrated uncertainty estimates.
By providing an automated tool for phonotrauma severity assessment, our work can enable large-scale studies of phonotrauma, ultimately leading to improved clinical understanding and patient care.
\end{abstract}

\paragraph*{Data and Code Availability}
We use a dataset of vocal fold images provided by Massachusetts General Hospital’s Center for Laryngeal Surgery and Voice Rehabilitation. For access, email kmatton@mit.edu to setup a data use agreement. Our code is available at: \href{https://github.com/kmatton/soft-ordinal-regression}{https://github.com/kmatton/soft-ordinal-regression}.

\paragraph*{Institutional Review Board (IRB)}
\vspace{-1mm}
Written informed consent was obtained from all subjects. The data used in this work came from two clinical research studies. The study protocols were reviewed and approved by an Institutional Review Board (IRB), protocol numbers 2011P002376 and 2008P000616.

\section{Introduction}
\label{sec:intro}

Approximately 20\% of adults in the United States experience a voice disorder at some point in their life, with even higher rates for singers and teachers \citep{huston2024prevalence}. One of the most common causes of voice disorders is detrimental habitual vocal behavior, referred to as vocal hyperfunction (VH). Phonotraumatic vocal hyperfunction (PVH) is a sub-type of VH characterized by vocal fold tissue damage (e.g., nodules, polyps). PVH can impair and even prevent normal vocal communication, leading to harmful social, economic, and personal consequences.

\begin{figure}
    \centering    \includegraphics[width=0.60\linewidth]{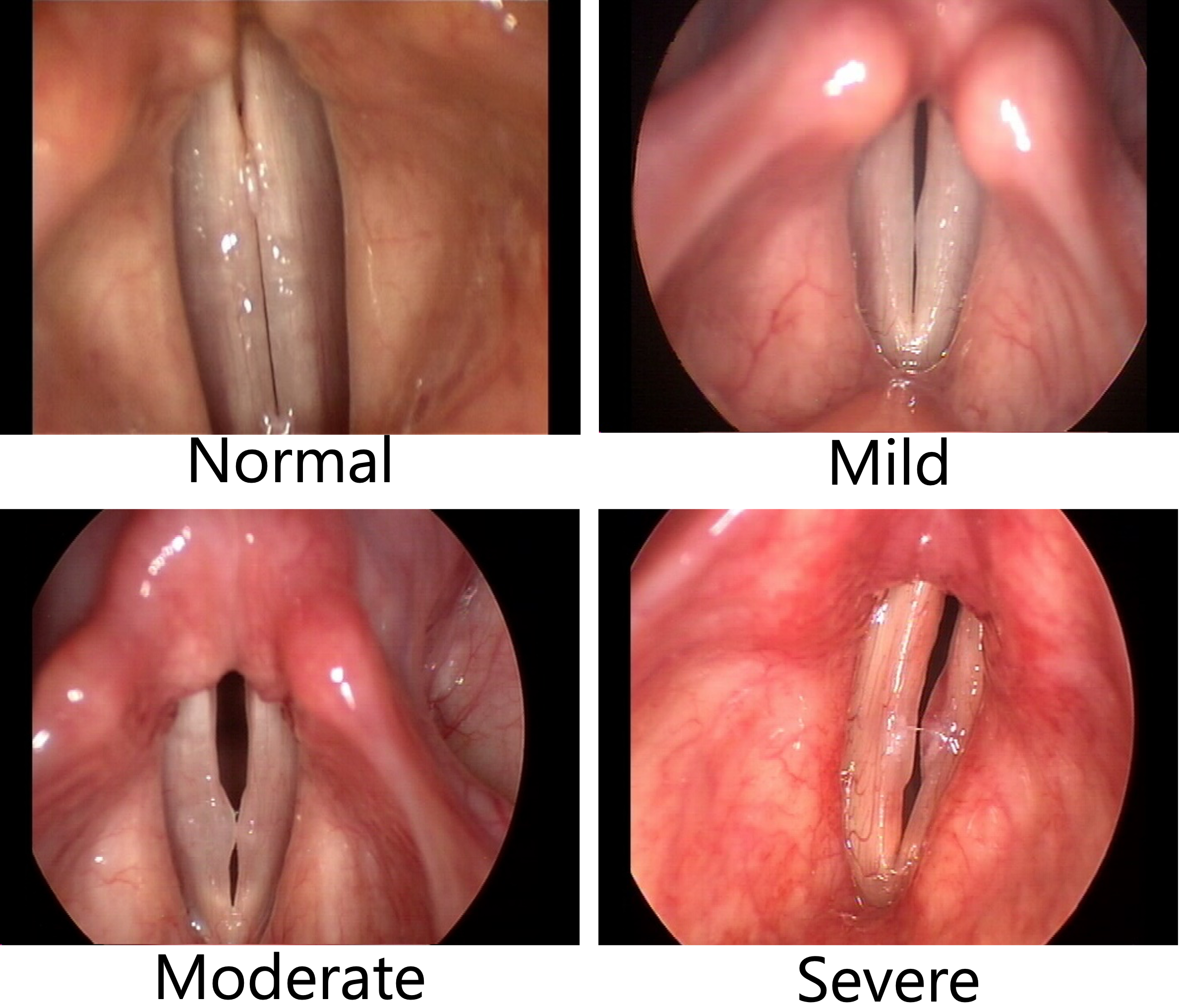}
    \caption{Images of vocal folds in the adducted (closed) position, showing varying levels of phonotrauma severity. Normal indicates healthy control.}
\label{fig:mgh_severity_examples}
\end{figure}

As shown in Figure~\ref{fig:mgh_severity_examples}, PVH can range from mild to severe. However, research studies often treat PVH as a binary variable, e.g., patients with PVH versus controls without PVH \citep{van2020differences, cortes2018ambulatory}. A more refined characterization of phonotrauma severity could help clarify the disorder’s etiological and pathophysiological mechanisms in earlier versus later stage tissue trauma \citep{cortes2018ambulatory}, as well as optimize prevention and treatment strategies for individual patients \citep{bequignon2013long}. Obtaining a large number of gold standard phonotrauma severity labels is difficult because it would rely on perceptual judgements by multiple laryngeal surgeons  \citep{van2023detecting}, which is expensive and time-consuming. Moreover, ratings are frequently inconsistent across even expert raters. Thus, there is a need for an automated phonotrauma severity assessment tool.

In this study, we present the first machine learning approach for automatically assessing phonotrauma severity from vocal fold images. We address two main modeling challenges. First, severity ratings have an inherent ordering (e.g., mild is less than moderate). Standard approaches to multi-class classification ignore ordinal structure, and therefore may be sub-optimal. Second, there is considerable label uncertainty. The differences between neighboring severity classes can be extremely subtle, with clinical experts often disagreeing on exactly where to draw category boundaries. On our set of vocal fold images with multiple ratings, the average pairwise inter-annotator agreement, measured by quadratic weighted kappa (QWK), is $0.61$, indicating only moderate agreement.

To handle label ordinality, we adopt a widely used ordinal regression framework that casts the severity classification problem as a series of binary classification sub-tasks \citep{frank2001simple, li2006ordinal}. We examine several state-of-the-art approaches that use this framework to train deep learning models with standard hard labels \citep{shi2023deep, cao2020rank, niu2016ordinal}. To account for uncertainty in severity ratings, we train models on soft labels that correspond to the distribution of annotator ratings. To achieve this, we propose a novel modification to standard ordinal regression loss functions that enables them to operate on soft labels. This simple yet important modification allows the model to learn from the disagreement among raters rather than discarding it.

We validate the utility of our approach through experiments on a dataset of vocal fold images collected from 214 patients, including 175 with varying phonotrauma severity levels and 39 healthy controls. We find that ordinal regression methods outperform standard multi-class classification in terms of predictive performance, and that training with soft labels improves the quality of uncertainty estimates. Our proposed soft ordinal regression method combines these two benefits: its performance approaches the mean inter-rater agreement between clinical expert pairs, and it produces the most well-calibrated uncertainty estimates among all methods considered.

Our main contributions are:
\begin{itemize}[leftmargin=*]
    \item We introduce phonotrauma severity assessment from images as a new machine learning task and demonstrate its feasibility, showing that model performance can approach that of experts. This opens new possibilities for objective, scalable severity assessment in clinical and research settings.
    \item We systematically evaluate multiple machine learning approaches for this new task, identifying ordinal regression and soft label training as effective techniques for addressing the task-specific challenges of label ordinality and label uncertainty.
    \item We adapt ordinal regression training objectives for the soft label setting and empirically demonstrate the utility of this new approach.
    \item We release the first dataset of vocal fold images with phontrauma severity annotations.
\end{itemize}
\section{Proposed Task: Phonotrauma Severity Assessment}
\subsection{Dataset}
\label{subsec:data}
\begin{table}[!t]
\centering
\small
\begin{tabular}{lcc}
\toprule
& \textbf{All} & \textbf{Multi-Rater} \\
\midrule
Total Count & 214 & 151 \\
Normal Count & 39 & 0 \\
Mild Count & 71 & 71 \\
Moderate Count & 68 & 68 \\
Severe Count & 36 & 12 \\
\midrule
Agreement (QWK) & $0.88^{1} \pm 0.01$ & $0.61 \pm 0.04$ \\
\bottomrule
\end{tabular}
\caption{Number of images per severity class in our dataset. The second column indicates the subset of images rated by three expert clinicians.}
\label{tab:data-stats}
\end{table}
\footnotetext[1]{Estimated upper-bound on QWK (cf. Section~\ref{subsec:data}).}

We collected a dataset of static vocal fold images from 214 subjects captured from videostroboscopy. For each subject, there are two images: one showing the vocal folds in the adducted (closed) position and one in the abducted (open) position. The adducted images were taken while the subject voiced at a high, soft pitch; this helps to show the full extent of phonotrauma. Example adducted images are in Figure~\ref{fig:mgh_severity_examples} and abducted images are in Appendix~\ref{app:abducted-images}. To assess phonotrauma severity, clinicians typically consider the presence and size of lesions (e.g., nodules and polyps), any signs of other trauma on the vocal fold tissue edges (e.g., redness, edema, scarring), and the effect of lesions on vocal fold closure during voicing (e.g., any gaps above or below the lesions).

There are 39 images from healthy controls. These images come from subjects with no history of a voice disorder who underwent a videostroboscopy screening with a speech-language pathologist to ensure that their larynx did not have any obvious abnormalities. The remaining 175 images come from patients with a phonotrauma diagnosis. For a majority of these images (151), three Laryngology Fellowship trained surgeon who specialize in the diagnosis and treatment of laryngeal disorders independently assessed the severity of phonotrauma as mild, moderate, or severe. We refer to this subset of the data as the \textit{Multi-Rater} subset. Because there were a lack of severe cases in the Multi-Rater subset, we added 24 additional severe cases by retrospectively reviewing approximately 200 patient records. To qualify for inclusion, three voice-specialized speech-language pathologists met in person and unanimously agreed on the images that represented severe phonotrauma.

The Multi-Rater subset exhibits considerable annotator disagreement: there is perfect consensus for only 49\% of images, with average pairwise inter-annotator agreement of $0.61$ QWK. To obtain a single ``hard'' label for each of image, we use the mode rating across annotators. Counts of each severity class (based on hard labels) are shown in Table~\ref{tab:data-stats}. To obtain an upper bound on inter-annotator agreement for the full dataset, we assume perfect agreement for all images without multiple ratings. The resulting upper bound is $0.88$ QWK, indicating high but imperfect agreement. 

While the images in our experiments have a standardized vocal fold position, they exhibit variability across other clinically relevant dimensions, including capture device (rigid vs. flexible endoscope), field of view, image angle, image quality, and color balance (brightness and saturation). In this work, we used only the adducted images because they are the most clinically informative. In future work, we plan to explore methods for incorporating the abducted images. 

\subsection{Problem Setup and Objective}
\label{subsec:problem-setup}

We denote the images in the dataset $\{\mathbf{x}_i\}_{i=1}^N$, where $N$ is the total number of images. Each image is a 3-channel RGB image, i.e., $\mathbf{x}_i \in \mathbb R^{H\times W\times 3}$. We let $\mathcal{Y} = \{1, \ldots, K-1, K\}$ denote the set of label categories, which are ordered as $1 \prec \ldots \prec K-1 \prec K$. For the phonotrauma severity task, $N=214$ and $K=4$, with $k = 1$ representing \textit{normal} and $k = 4$ representing \textit{severe}. We let $p_i^k$ denote the empirical probability that image $\mathbf{x}_i$ is labeled as severity rating $k$ (i.e., the fraction of annotators that gave this rating). The \textit{soft} label for $\mathbf{x}_i$ is $\mathbf{y}_i = [p_i^1, \ldots p_i^K]$. The \textit{hard} label is the mode rating $y_i := \operatorname{arg\,max}_k \mathbf{y}_i$.

Our goal is to learn a function $f: R^{H\times W\times 3} \rightarrow [0, 1]^K$ that maps images to a probability distribution over severity categories, where $\sum_{k=1}^K f(\mathbf{x}_i)_k = 1$. Given the predicted probabilities, we  obtain a single hard prediction as $\hat{y}_i := \operatorname{arg\,max}_k f(\mathbf{x}_i)_k$. Our primary goal is to maximize performance in predicting hard severity labels on held-out test data (i.e., $\hat{y}_i$ is close to $y_i$). Our secondary goal is to obtain well-calibrated uncertainty estimates, such that the maximum predicted probability reflects the true probability that the prediction is correct. This is important to allow practitioners to assess the reliability of automated predictions.

\section{Background: Ordinal Regression with Hard Labels}\label{sec:background}
A widely-used approach to ordinal regression is to decompose the problem into a series of binary classification tasks \citep{frank2001simple, li2006ordinal}. We review two state-of-the-art deep learning methods that are based on this framework and are designed to work with hard labels. In Section~\ref{sec:proposed-approach}, we extend them to enable soft label training.

\textbf{(1) OR-CNN} \citep{niu2016ordinal} casts the problem as $K-1$ binary classification sub-tasks, where the $k$-th task is to predict whether the label exceeds rank $k$. To perform these tasks, OR-CNN trains a single neural network with a shared encoder $g(\cdot)$ and task-specific classification heads $h_k(\cdot)$. The output of the model for task $k$ is $h_k(g(\mathbf{x}_i))$. This represents the predicted probability that label $y_i$ exceeds rank $k$, i.e., $h_k(g(\mathbf{x}_i)) = \hat{P}(y_i > k)$. The method was originally developed for image data and uses a convolutional neural network (CNN) as the encoder $g(\cdot)$. 

To optimize model parameters, OR-CNN uses a loss function corresponding to the sum of task-specific losses, where each task-specific loss is a standard binary cross entropy (BCE) loss. Let $y_i^{>k}$ be a task-specific label indicating whether label $y_i$ exceeds rank $k$, i.e., $y_i^{>k} = \mathds{1}_{y_i > k}$. The aggregate loss function is:
\begin{align}
\mathcal{L}
&= -\!\!\sum_{i=1}^N \sum_{k=1}^{K-1} \Bigl[
    y_i^{>k} \log h_k(g(\mathbf{x}_i))  \notag\\
&\qquad\qquad +\; (1 - y_i^{>k}) \log\bigl(1 - h_k(g(\mathbf{x}_i))\bigr)
\Bigr]
\label{eqn:or-loss}
\end{align}

To obtain a prediction of the rank for an image $\mathbf{x}_i$, OR-CNN counts the number of sub-tasks with predicted probability greater than $0.5$, i.e., $\hat{y}_i= 1 + \sum_{k=1}^{K-1} \mathds{1}_{h_k(g(\mathbf{x}_i)) > 0.5}$. One known issue with OR-CNN is that it can yield inconsistent probability estimates across tasks (e.g., $\hat{P}(y_i > 1) < \hat{P}(y_i > 2)$).

\textbf{(2) CORAL} \citep{cao2020rank} was introduced to resolve the rank-inconsistency problem of OR-CNN. It follows the same approach as OR-CNN, with one modification: CORAL shares output layer weights across tasks and only includes task-specific bias terms. For each task $k$, $h_k(g(\mathbf{x}_i)) = \sigma\bigl(g(\mathbf{x}_i) + b_k\bigr)$. If the bias terms are non-increasing, i.e., $b_1 \geq b_2 \geq \ldots b_{K-1}$, then the predicted probabilities will also be non-decreasing, achieving rank consistency.

\section{Proposed Approach: Ordinal Regression with Soft Labels}\label{sec:proposed-approach}
\begin{figure*}
    \centering
    \includegraphics[width=0.9\linewidth]{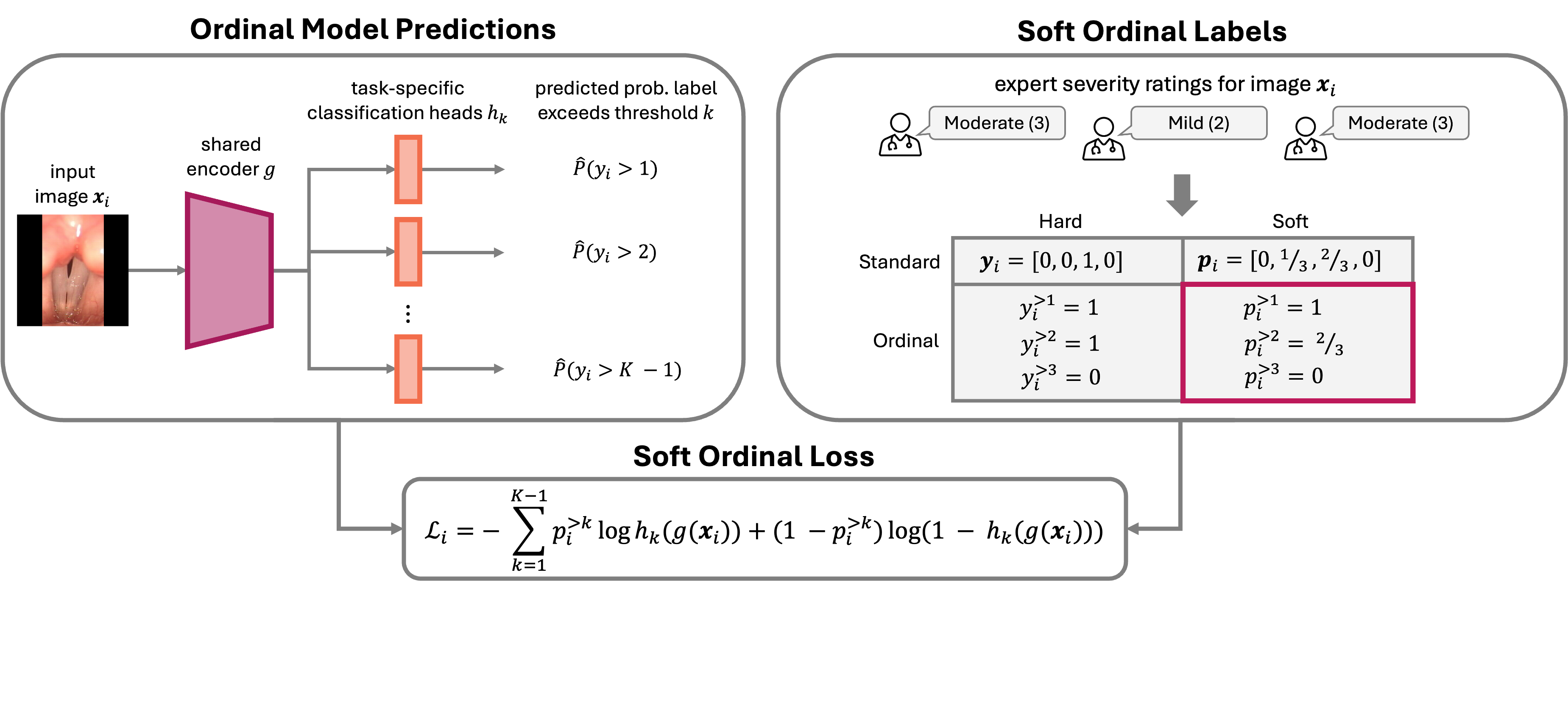}
    \caption{Overview of our proposed soft ordinal regression approach. \textit{(1) Ordinal Model Predictions:} as in prior work \citep{niu2016ordinal, cao2020rank}, we train a model to perform $K-1$ tasks, where the $k$-th task is to predict whether the label exceeds rank $k$. \textit{(2) Soft Ordinal Labels:} we  form soft labels corresponding to the empirical probability that the label exceeds rank $k$. \textit{(3) Soft Ordinal Loss:} we sum over task-specific \textit{weighted} binary cross entropy loss terms; the weights correspond to soft label probabilities from part (2). }
    \label{fig:method-diagram}
\end{figure*}

The ordinal regression methods in Section~\ref{sec:background} are designed to work with standard (hard) labels. In this section, we propose a novel modification to the loss function that enables training with soft labels.

We first compute task-specific soft labels for each example $i$. For task $k$, we use the empirical probability that example $i$ has rank greater than $k$, based on its distribution of annotator ratings. Specifically, for example $i$ with soft label $[p_i^1, \ldots, p_i^K]$ we compute $p_i^{>k} = \sum_{i = k + 1}^K p_i^k$. Given these soft labels, we optimize the loss function:
\begin{align}
\mathcal{L}
&= -\!\!\sum_{i=1}^N \sum_{k=1}^{K-1} \Bigl[
    p_i^{>k} \log h_k(g(\mathbf{x}_i))  \notag\\
&\qquad\qquad +\; (1 - p_i^{>k}) \log\bigl(1 - h_k(g(\mathbf{x}_i))\bigr)
\Bigr]
\label{eqn:or-soft-loss}
\end{align}
This is the same as Equation~\ref{eqn:or-loss} except each term is \textit{weighted} BCE, where the weights are computed based on each example's empirical label distribution. We refer to our proposed variants of OR-CNN and CORAL that use this loss function as \textbf{OR-Soft} and \textbf{CORAL-Soft}. Our method is depicted in Figure~\ref{fig:method-diagram}.
\section{Experiments}
\label{sec:exp}

\subsection{Experimental Settings}\label{subsec:exp-settings}
 For the encoder network $g(\cdot)$, we use a ResNet-50 \citep{he2016deep} pre-trained on ImageNet \citep{deng2009imagenet}. We use a linear layer for each task-specific output head $h_k(\cdot)$. To obtain robust performance estimates with our small dataset, we employ five-fold cross-validation with stratified sampling to ensure balanced severity class representation across folds. We report the mean and standard deviation of performance across folds. Further details on experimental settings, including data pre-processing, data augmentations, and hyperparameters, are in Appendix~\ref{app:impl-details}.

\subsection{Baselines}\label{subsec:baselines}
We compare our proposed method to standard multi-class classification and ordinal regression baselines. For all baselines, we use the experimental settings described in Section~\ref{subsec:exp-settings}.

\paragraph{Standard Multi-Class Classification.} We consider two standard methods for multi-class classification that treat the labels as unordered categories. (1) CE: we train a model using hard labels and multi-class cross entropy (CE) loss. (2) CE-soft: we use CE loss with soft labels.

\paragraph{Ordinal Regression.} In addition to OR-CNN and CORAL (cf. Section~\ref{sec:background}), we consider two methods. The first method, CORN \citep{shi2023deep}, casts ordinal regression as a series of binary sub-tasks, like OR-CNN and CORAL. However, CORN proposes slightly different tasks: the $k$-th task is to predict whether the label exceeds rank $k$, \textit{conditional} on the label being at least rank $k$. This reformulation allows CORN to achieve rank-consistency without the weight-sharing constraint of CORAL. CORN trains each task on a subset of examples selected based on their hard label, making it non-trivial to extend to soft labels. The second method, SORD \citep{diaz2019soft}, creates synthetic soft labels by smoothing around the true hard label and then trains with standard CE loss. To create smoothed labels, SORD computes the distance of each class from the ground-truth class, placing higher probability on nearby classes. We examine SORD with two distance metrics: (1) absolute error (AE) and (2) squared error (SE).

\subsection{Evaluation Metrics}\label{subsec:eval}
\paragraph{Ordinal-Aware Metrics.} We use two metrics that penalize distant errors more than adjacent ones: (1) Mean absolute error (MAE), which measures error with a linear penalty, and (2) Quadratic Weighted Kappa (QWK), which measures agreement while applying quadratic penalties to errors. 

\paragraph{Uncertainty-Weighted (UW) Metrics.} When computing each metric, we treat the mode clinical rating as the gold standard label. However, given that there is considerable annotator disagreement, some labels are more reliable than others. To account for this, we consider uncertainty-weighted variants of each metric, in which we give a greater weight to examples with greater agreement. Let $w_i$ be the proportion of annotators that selected the mode rating for example $i$ and let $m(\hat{y}_i, y_i)$ be the metric value (e.g., absolute error) for example $i$. We compute the uncertainty-weighted (UW) metric as: $\frac{1}{\sum_{i=1}^N w_i} {\sum_{i=1}^N} w_i \cdot m(\hat{y}_i, y_i)$, where $N$ is the number of examples in the dataset.

\paragraph{Any-Rater (AR) Accuracy.} In addition to standard accuracy (agreement with the mode label), we report any-rater (AR) accuracy, which measures the rate at which model predictions match at least one clinical annotator. This metric reflects the idea that agreement with any clinical expert constitutes a plausible prediction given the inherent subjectivity of severity assessment.

\paragraph{Uncertainty Quantification Metric.} To evaluate the quality of uncertainty estimates, we compute Expected Calibration Error (ECE), which measures how well predicted confidence aligns with true accuracy. We treat the predicted probability of the model's top prediction as its confidence and compute the expected true accuracy for examples at each confidence level by using the soft label probabilities.

\paragraph{Composite Metric.} As a metric that captures both the quality of model predictions and of uncertainty estimates, we use area under the risk-coverage curve (AURC). AURC is designed to reflect the clinically practical scenario of using a machine learning model for \textit{selective classification}, i.e., the model makes predictions when its confidence exceeds some threshold and otherwise abstains. A risk-coverage curve is generated by plotting the proportion of data points the model makes predictions for (i.e., coverage) on the x-axis and the error rate for those predictions on the y-axis.  We use the uncertainty-weighted (UW) error rate, which is 1 - Accuracy (UW).

\begin{table*}[!t]
\centering
\small
\begin{tabular}{lcccccc}
\toprule
Method & MAE (UW) & QWK (UW) & Accuracy (UW) & Accuracy (AR) & ECE & AURC \\
 \midrule
 CE &  0.29 $\pm$ 0.09 &  0.82 $\pm$ 0.09 &  0.74 $\pm$ 0.06 &  0.86 $\pm$ 0.05 & 0.17 $\pm$ 0.04 & 0.15 $\pm$ 0.05 \\
 CE-Soft & 0.32 $\pm$ 0.09 &  0.80 $\pm$ 0.07 &  0.72 $\pm$ 0.07 &  0.84 $\pm$ 0.06 &  \underline{0.10 $\pm$ 0.03} & 0.12 $\pm$ 0.04  \\
CORN &  \textbf{0.23 $\pm$ 0.08} &  \textbf{0.86 $\pm$ 0.07} &  \textbf{0.79 $\pm$ 0.07} &  \textbf{0.89 $\pm$ 0.04} & 0.16 $\pm$ 0.03 & 0.11 $\pm$ 0.06 \\
SORD-AE & 0.28 $\pm$ 0.07 & 0.82 $\pm$ 0.08 & 0.73 $\pm$ 0.07 & 0.86 $\pm$ 0.04 & 0.26 $\pm$ 0.06 & 0.12 $\pm$ 0.04 \\
SORD-SE  & 0.28 $\pm$ 0.07 & 0.82 $\pm$ 0.08 & 0.73 $\pm$ 0.06 & 0.85 $\pm$ 0.02 & 0.20 $\pm$ 0.04 & 0.13 $\pm$ 0.03 \\
CORAL & 0.51 $\pm$ 0.12 & 0.74 $\pm$ 0.08 & 0.55 $\pm$ 0.08 & 0.66 $\pm$ 0.11 & 0.29 $\pm$ 0.05 & 0.32 $\pm$ 0.04\\
 CORAL-Soft & 0.47 $\pm$ 0.09 & 0.75 $\pm$ 0.06 & 0.58 $\pm$ 0.05 & 0.68 $\pm$ 0.06 & 0.26 $\pm$ 0.05 & 0.31 $\pm$ 0.04 \\
 OR-CNN &  \underline{0.26 $\pm$ 0.08} &  0.83 $\pm$ 0.08 &  \underline{0.76 $\pm$ 0.06} &  \underline{0.87 $\pm$ 0.05} & 0.15 $\pm$ 0.04 & \underline{0.10 $\pm$ 0.02} \\
 OR-Soft &  \underline{0.26 $\pm$ 0.09} &  \underline{0.85 $\pm$ 0.06} &  0.75 $\pm$ 0.08 &  0.86 $\pm$ 0.04 &  \textbf{0.10 $\pm$ 0.06} & \textbf{0.09 $\pm$ 0.03} \\
\bottomrule
\end{tabular}
\caption{Performance (mean $\pm$ standard deviation) across five folds. The method with the best average performance is in \textbf{bold}, second-best is \underline{underlined}.  UW = Uncertainty-Weighted, AR = Any-Rater Accuracy (see Section~\ref{subsec:eval}). CORN \citep{shi2023deep} achieves the best predictive performance, whereas our proposed OR-Soft  method achieves the best balance of high predictive performance and low calibration error.}
\label{tab:methods_soft_hard}
\end{table*}

\subsection{Results}
\paragraph{Main Results.} As shown in Table~\ref{tab:methods_soft_hard}, CORN performs the best in terms of mean predictive performance: it obtains the lowest MAE ($0.23$), highest QWK ($0.86$), and highest accuracy scores ($0.79$ for predicting the
mode label and $0.89$ for predicting any expert label). Both OR-CNN and our proposed OR-Soft method also exhibit strong predictive performance, reaching scores that are similar to CORN. OR-Soft obtains the second lowest MAE ($0.26$), second highest QWK ($0.85$), and relatively high accuracy scores ($0.75$ for the mode label and $0.86$ for any expert label). Using a paired one-sided t-test, we find that there is not a statistically significant difference between the predictive performance of CORN and OR-Soft (details in Appendix~\ref{app:significant-tests}). Both CORN and OR-Soft obtain QWK scores that approach the mean pairwise agreement between expert annotators, which is upper-bounded by $0.88$ QWK (cf. Section~\ref{subsec:data}). 

Examining calibration performance, we find that methods trained with (natural) soft labels consistently produce more accurate uncertainty estimates than their hard-label variants. OR-Soft and CE-Soft tie for the lowest mean ECE of $0.10$. Despite CORN's superior predictive performance, the method has relatively poor calibration performance (mean ECE of $0.16$). OR-Soft obtains a statistically significant improvement in ECE over CORN, as discussed in Appendix~\ref{app:significant-tests}. Training with synthetic soft labels does not offer the same benefit as training with soft labels obtained from multi-rater data: both versions of SORD obtain relatively high mean ECE values.

 Among all methods, OR-Soft obtains the best balance between predictive performance and uncertainty estimation. It achieves close to the best MAE and QWK, while obtaining the lowest ECE. It also obtains the lowest AURC, indicating that it can both produce accurate predictions and use its uncertainty estimates to discriminate between correct and incorrect predictions.

The CORAL-based methods perform worse than the other methods in terms of both predictive performance and ECE. We hypothesize that this may because the weight-sharing constraint of CORAL is too restrictive for our phonotrauma severity task; for example, the features distinguishing normal from mild cases may differ substantially from those distinguishing moderate from severe cases.

In Table~\ref{tab:distribution_metrics} (Appendix~\ref{app:additional-results}), we present results for two metrics that quantify the distance between the true and predicted label distributions: cross entropy and Brier score. We find that OR-Soft performs the best in terms of these metrics. We present additional predictive performance metrics, including AUROC and Spearman correlation, in Table~\ref{tab:coverage_auc_spearman} (Appendix~\ref{app:additional-results}); we find that the relative performance across methods is similar to the predictive performance metrics we present in the main text.

\paragraph{Confusion Matrix Analysis.} \begin{figure}
    \centering
    \includegraphics[width=1\linewidth]{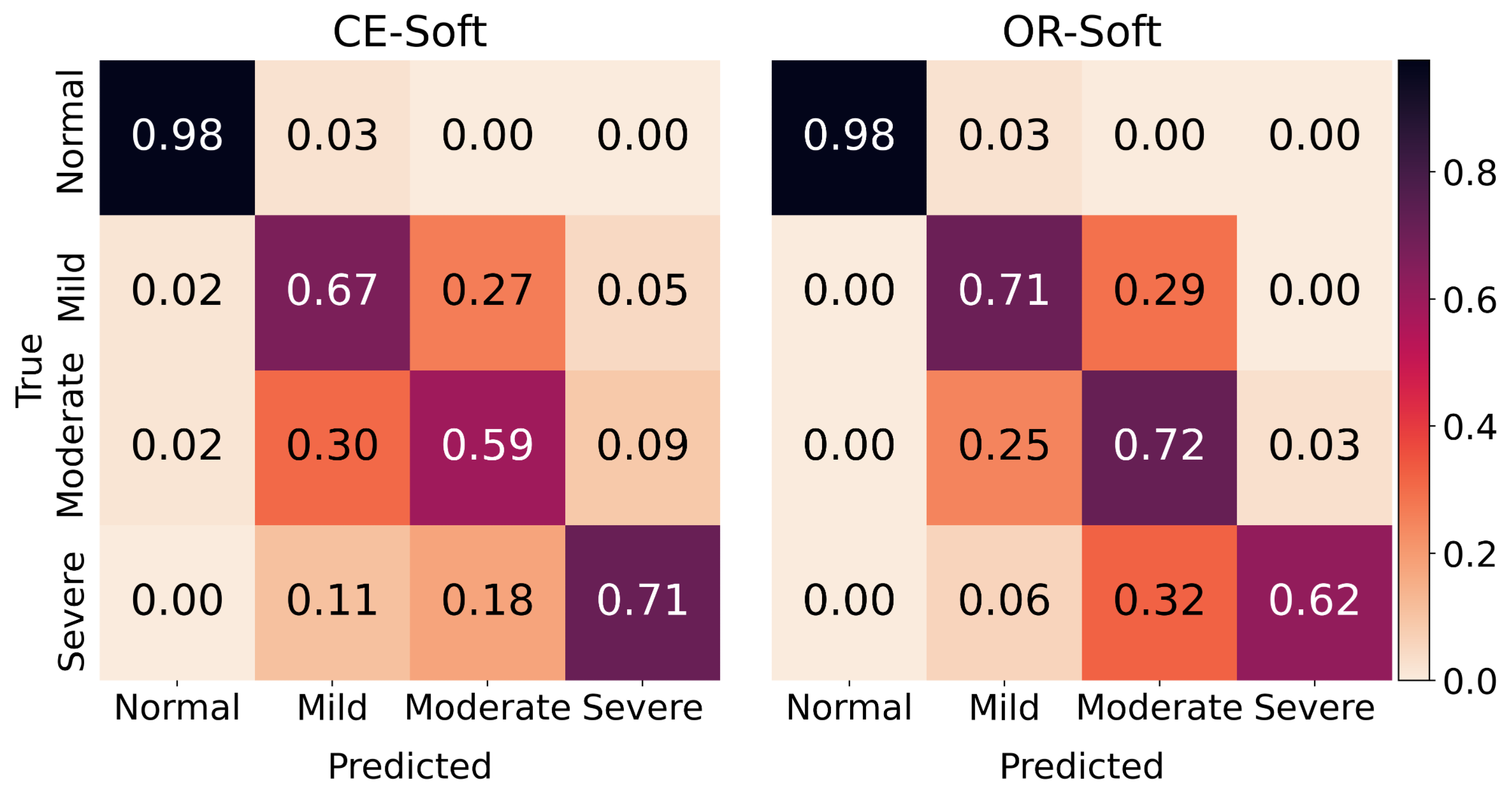}
    \caption{Confusion matrices for CE-Soft and OR-Soft. Both methods have high accuracy in discriminating between normal and non-normal cases. OR-Soft makes fewer off-by-two errors than CE-Soft.}
    \label{fig:cm}
\end{figure} To understand how ordinal regression affects predictions, we compare confusion matrices for OR-Soft and CE-Soft in Figure~\ref{fig:cm}. Because of space constraints, we show the average row-normalized confusion matrix across folds here and include a version with standard deviations in Figure~\ref{fig:cm-with-std} (Appendix~\ref{app:additional-results}). Both methods excel at distinguishing normal from abnormal cases, achieving near-perfect recall of normal cases ($0.98$). OR-soft never misclassifies pathological cases as normal, whereas CE-Soft does so rarely (in $2\%$ of mild and moderate cases). The methods differ in their error patterns for intermediate severity levels. CE-soft struggles most with moderate cases (recall = $0.59$), frequently misclassifying them as mild. In contrast, OR-Soft struggles most with severe cases (recall = $0.62$), which it typically misclassifies as moderate. OR-Soft makes fewer off-by-two-category errors compared to CE-Soft. This result confirms our intuition that accounting for ordinal structure can help to prevent clinically implausible errors.

\paragraph{Calibration Curve Analysis.}
\begin{figure}
    \centering
    \includegraphics[width=1\linewidth]{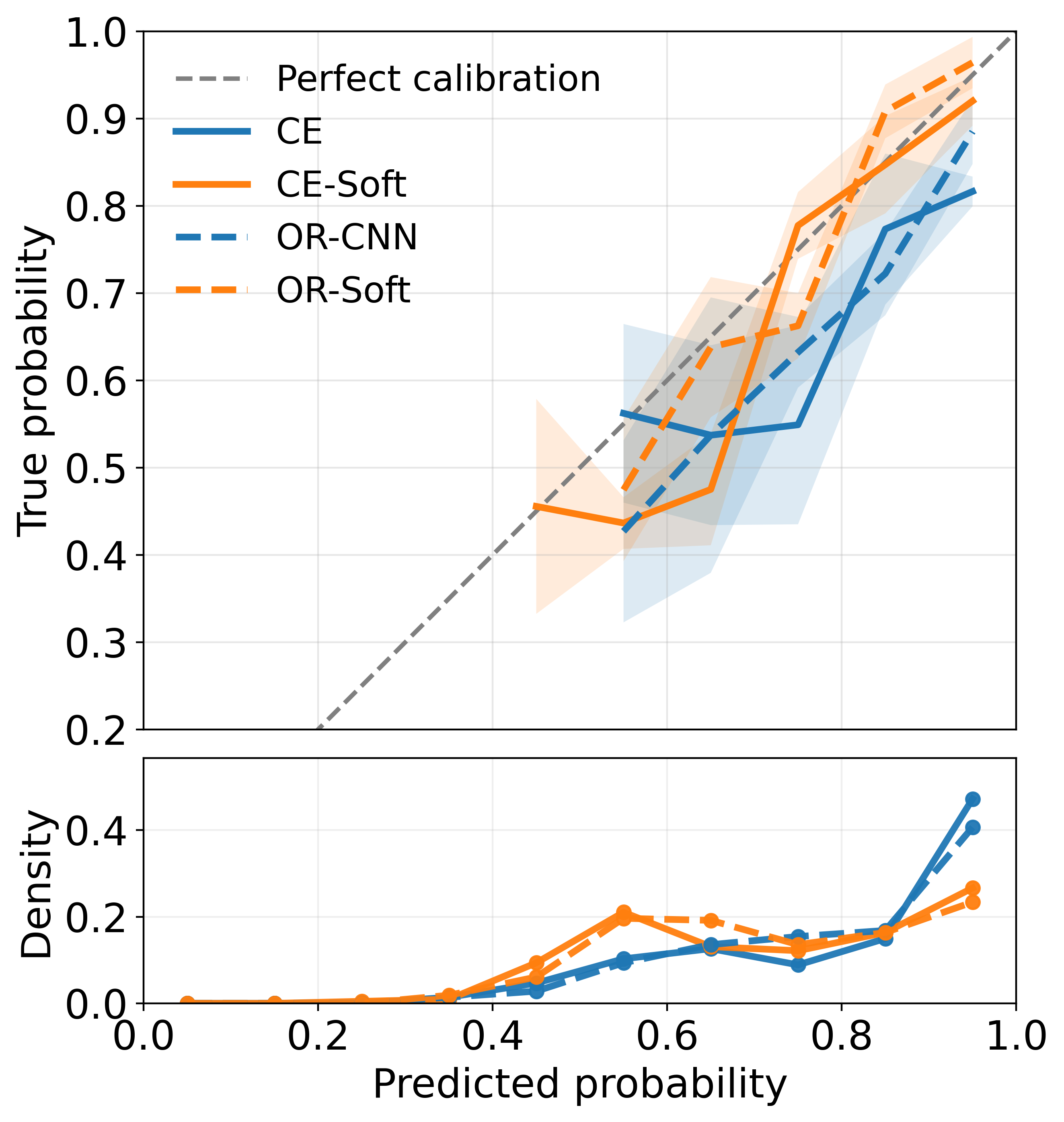}
    \caption{\textit{Top:} calibration curves for hard methods (blue) and their soft variants (orange). The orange curves are consistently closer to the gray line (perfect calibration) than the blue curves. \textit{Bottom:} density of predicted probabilities for each method. }
    \label{fig:calibration}
\end{figure} To understand the impact of soft label training on uncertainty estimation, we examine the calibration curves for standard methods and their soft variants in Figure~\ref{fig:calibration}. The x-axis represents the model's predicted confidence (maximum predicted probability across classes), and the y-axis represents the true expected accuracy (given multi-annotator ratings) at that confidence level. Perfect calibration follows the dashed gray diagonal. The soft methods (orange) demonstrate superior calibration compared to hard methods (blue). The hard methods exhibit systematic over-confidence, particularly at high predicted probabilities, as shown by their curves falling consistently below the diagonal line. This over-confidence is a well-known issue with hard label training \citep{wang2023calibration}. OR-Soft shows the best overall calibration; its curve is closer to the diagonal than all other methods for most confidence levels.

\paragraph{Risk-Coverage Curve Analysis.}
\begin{figure}
    \centering
    \includegraphics[width=1\linewidth]{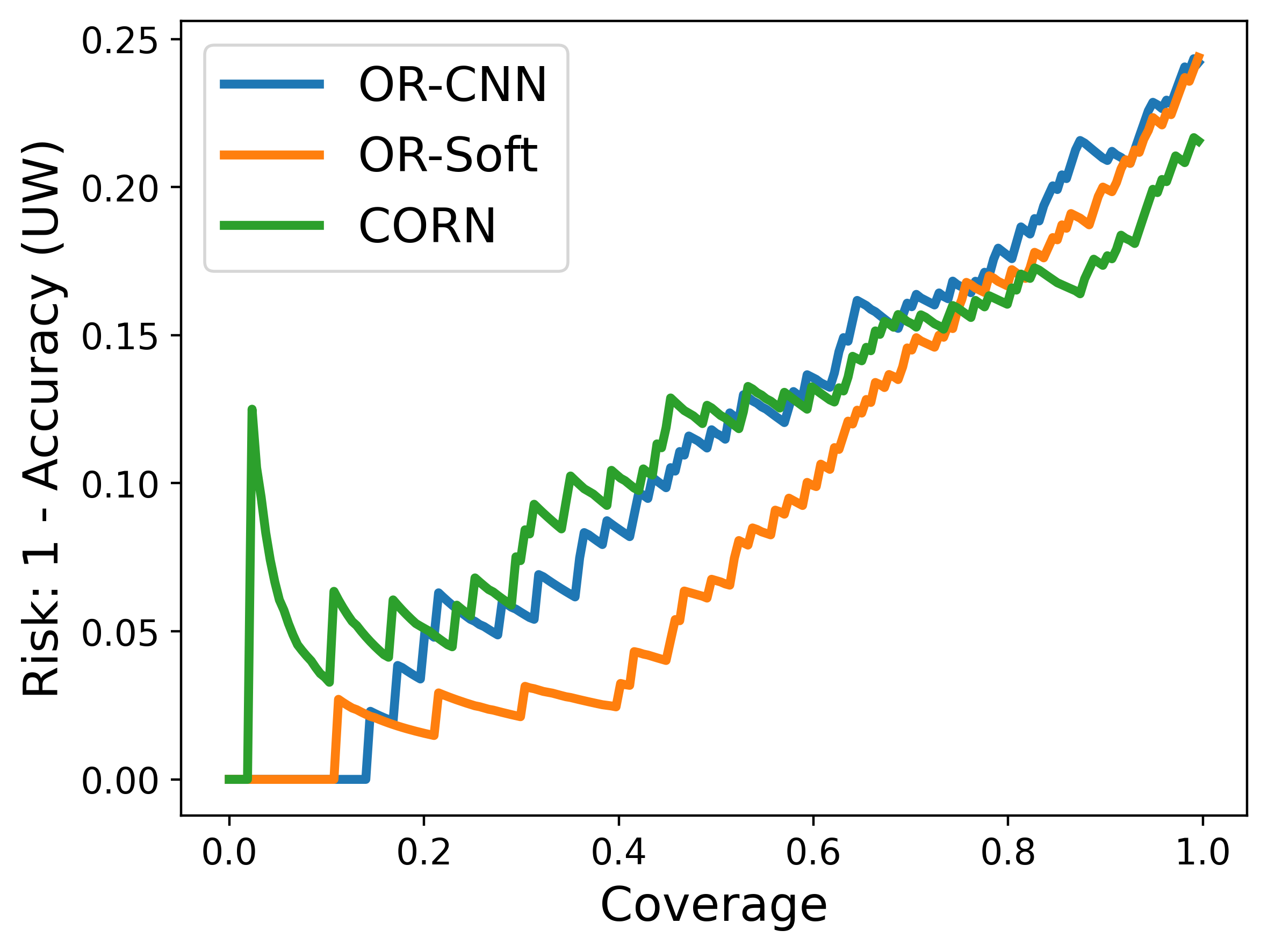}
    \caption{Risk-coverage curves for the three methods with the lowest AURC. OR-Soft has the lowest risk for low to moderate coverage rates, whereas CORN has the lowest risk for high coverage rates.}
    \label{fig:risk-coverage}
\end{figure}
In Figure~\ref{fig:risk-coverage}, we present risk-coverage curves for the three methods with the lowest AURC: OR-Soft, OR-CNN, and CORN. For ease of visualization, we generate a single curve per method by combining the predictions and labels for examples across each test fold. We present curves for individual folds in Appendix~\ref{app:additional-results}. OR-Soft (in orange) obtains the lowest error rate (y-axis) for most coverage values (x-axis), including the interval $[0.2, 0.75]$. CORN obtains the lowest error rate for high coverage rates (around $0.8$ and above). These results indicate that the preferred method depends on the desired risk-coverage tradeoff in the deployment setting. OR-Soft's superior performance at lower coverage rates makes it well-suited for settings where uncertain cases can be deferred to specialists. CORN may be preferred when near-complete automation is required and slightly higher risk is acceptable.

\section{Related Work}
\label{sec:related-work}
\paragraph{Classification of Vocal Fold Images.}
To the best of our knowledge, there is no existing work on classifying the severity of phonotrauma from vocal fold images. 
Existing machine learning approaches to vocal fold image analysis have focused on other related tasks, including binary classification of vocal fold normality \citep{cho2022comparison, tran2023support}, detecting the presence of lesions \citep{larsen2023comparison, yao2024deep}, determining if a lesion is benign or malignant \citep{bur2023interpretable, dao2024improving}, and classifying lesions types \citep{verikas2006towards, ren2020automatic, zhao2022vocal, kim2023convolutional}. While early work relied on hand-crafted features and classical machine learning algorithms, recent approaches have adopted deep convolutional neural networks (CNNs) \citep{ren2020automatic, zhao2022vocal, larsen2023comparison, yao2024deep}, often leveraging transfer learning from models pretrained on the ImageNet dataset \citep{deng2009imagenet}.

\paragraph{Ordinal Regression.}
The statistical literature on ordinal regression is extensive; we refer the reader to \citet{tutz2022ordinal} for a review. Here, we focus on ordinal regression methods that have been adapted for deep learning. There are three main types of approaches. The first type of approach frames ordinal regression as a series of binary sub-tasks \citep{frank2001simple, liu2020unimodal}. We described this approach in Section~\ref{sec:background} and adapt it as part of our proposed soft ordinal regression approach (see Section~\ref{sec:proposed-approach}). CORN \citep{shi2023deep}, a baseline described in Section~\ref{subsec:baselines}, falls in this category. The second type of approach generates \textit{synthetic} soft labels, which are used to encode that near-by classes are more similar than distant ones. SORD \citep{diaz2019soft}, described in Section~\ref{subsec:baselines}, exemplifies this method type. Whereas these methods generate synthetic soft labels as a way of encoding ordinal structure, we incorporate natural soft labels (from multi-annotator ratings) as a way of capturing label uncertainty. Finally, a third style of approach is based on constraining models to predict uni-modal distributions \citep{da2005classification, beckham2017unimodal, liu2020unimodal, albuquerque2021ordinal, yamasaki2022unimodal, cardoso2025unimodal}. These methods are designed for hard label supervision rather than soft labels.

\paragraph{Learning with Multi-Rater Annotations.}
In medical imaging, annotations are often provided by multiple expert raters, leading to the challenge of how to incorporate these labels during training. The simplest approaches use consensus strategies such as majority voting \citep{nabulsi2021deep}, but these discard information about uncertainty and variability across annotators. Label distribution learning (LDL) provides a framework in which each data instance is associated with a label distribution \citep{geng2016label}. Label distributions can be formed in multiple ways, including by treating votes from multiple raters as a fractional distribution over classes \citep{wang2019classification}. Most LDL approaches train models with distribution-matching losses that are not intended for ordinal data (e.g., KL-Divergence) \citep{gao2017deep, wang2019classification, wang2021re, li2020deep}. More recently, \citet{wen2023ordinal} proposed Ordinal LDL: training using the cumulative distribution function (CDF) and order-sensitive distributional loss functions. However, in many clinical problems such as ours, there are very few raters, leading to label distribution estimates that may be too coarse to benefit from a fully distributional approach. Hence, we focus on adapting hard-label ordinal regression methods to incorporate soft-label supervision.

\section{Summary and Discussion}
\vspace{-1.5mm}
Current approaches to phonotrauma severity assessment rely on perceptual judgments by multiple laryngeal surgeons, which are expensive and time-consuming to obtain. We present the first study that demonstrates the feasibility of automating phonotrauma severity assessment using machine learning. We systematically examine machine learning approaches for phonotrauma severity assessment that are tailored to the task-specific challenges of label ordinality and label uncertainty. We propose a novel approach, soft ordinal regression (OR-Soft), for jointly handling both challenges. We demonstrate that our proposed approach achieves a strong balance of predictive performance and uncertainty estimation compared to baselines.

\paragraph{Accuracy-Uncertainty Tradeoff.}
Among the methods examined in this study, we observed a trade-off between predictive accuracy and uncertainty estimation quality (see Appendix~\ref{app:tradeoff} for a visualization). CORN \citep{shi2023deep} achieved the best predictive performance, but it produced poorly calibrated uncertainty estimates. Conversely, OR-Soft obtained well-calibrated uncertainty estimates at the cost of slightly lower predictive accuracy. This raises the question of whether this trade-off is intrinsic or if it could be resolved through methodological refinement. Across all methods where we evaluated both hard and soft label variants, we found that soft variants achieved better-calibrated uncertainty estimates while maintaining comparable predictive performance. This suggests that developing a soft variant of CORN could potentially resolve the accuracy-uncertainty trade-off observed in our experiments. The creation of a soft-CORN approach is non-trivial because CORN relies on hard labels to partition the data and achieve rank-consistency.

\paragraph{Limitations.}
A primary limitation of our work is that we evaluate on a single dataset of limited size. Since we introduce both a novel clinical machine learning task and the first dataset for addressing it, no other suitable dataset exists for evaluation. This constraint limits our ability to make claims about the generalizability of our approach. In future work, we plan to collect additional datasets to validate whether our findings transfer to other contexts. 

In the Multi-Rater subset of the data, the severity ratings are provided by three laryngeal surgeons with diagnostic authority, ensuring clinical validity. However, annotations from three experts may not fully capture the range of diagnostic variation across the broader community of laryngeal specialists. Additionally, the severity class distributions differ substantially between the Multi-Rater subset and the remainder of the dataset (see Table~\ref{tab:data-stats}), which could influence our experimental results and limit generalizability to other label acquisition procedures.

To address these concerns, we collected annotations from three additional laryngeal surgeons for all subjects in the dataset. We acquired these labels after the submission deadline and have conducted a preliminary analysis. The results, shown in Appendix~\ref{app:full-data}, largely align with the findings in the main text. OR-Soft achieves the best balance between predictive performance and uncertainty estimation, and the soft label methods have superior uncertainty calibration compared to their hard label variants. One difference is that OR-Soft performed slightly better than CORN on this expanded label set in terms of MAE and QWK; however, we found that this difference was not statistically significant, consistent with our main findings.

\paragraph{Broader Impact.}
By paving the way for automated phonotrauma severity assessment, our work can help to enable population-scale analysis of phonotraumatic vocal hyperfunction. Such studies are critical for improving clinical understanding of the disorder, which can ultimately inform more effective treatment strategies and improve patient outcomes. In addition, by releasing our dataset, we hope to stimulate broader participation in research on phonotrauma severity assessment.

\acks{This research was supported by the National Institutes of Health (NIH) National Institute on Deafness and Other Communication Disorders (Grants P50 DC015446 and R33 DC011588), the NIH National Institute of Mental Health (Grants UM1MH130981, R01 MH123195, R01 MH121885, 1RF1MH123195), the Matthew Kerr Fellowship Fund, and Quanta Computer Inc. We would like to thank Ahmed Yousef for his helpful feedback and discussions. }

\bibliography{jmlr-sample}

\appendix

\section{Additional Vocal Fold Images}\label{app:abducted-images}

\begin{figure}[h]
    \centering
    \includegraphics[width=1\linewidth]{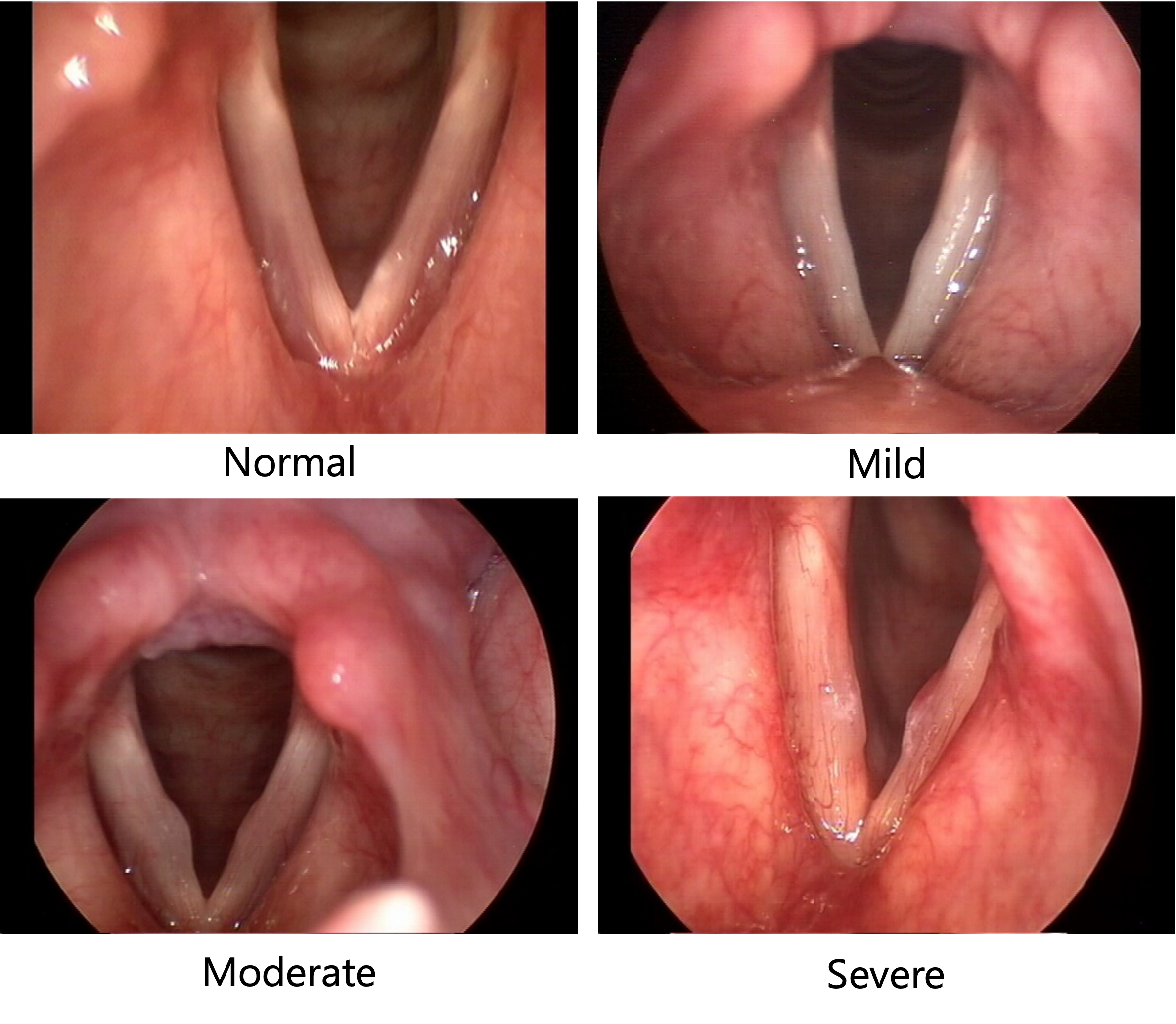}
    \caption{Images of vocal folds in the abducted (open) position, showing varying levels of phonotrauma severity. Normal indicates healthy control.}
    \label{fig:mgh_severity_examples_abducted}
\end{figure}

Figure~\ref{fig:mgh_severity_examples} in the main text shows examples of vocal folds in the adducted (closed) position. In Figure~\ref{fig:mgh_severity_examples_abducted}, we provide examples from our dataset showing vocal folds in the abducted (open) position.

\section{Experimental Details}\label{app:impl-details}

\subsection{Data Pre-Processing}
The images in our dataset vary in size, orientation, and color. We standardize them by applying the following pre-processing steps: (1) \textit{center cropping} along the width dimension by a factor $0.9$ to remove parts of the image not relevant to severity prediction (e.g., anatomy surrounding the vocal folds), (2) \textit{resizing with padding} to a fixed target size ($554$ by $544$) while preserving the original aspect ratio, and (3) \textit{color normalization} by min-max scaling.

\subsection{Data Augmentations} During model training, we apply data augmentations to encourage robustness to aspects of the image that are not relevant to the severity prediction task. The augmentations we use include: cropping, rotation, horizontal flipping, adjusting brightness and contrast, Gaussian noise, Gaussian blurring, and gamma correction. In addition, we create a custom augmentation to simulate the black circular borders that appear in some but not all images (e.g., as in Figure~\ref{fig:mgh_severity_examples}).

\subsection{Hyperparameters}
We train all models for 1000 epochs. We use a batch size of 16 and a learning rate of $0.00001$. We use the Adam Optimizer.

\subsection{Model Selection}
For each experiment, we split the training set into training examples ($80\%$) and validation examples ($20\%$). We select the best model from the 1000 training epoch based on the validation performance. We use MAE with uncertainty-weighting (cf.~\ref{subsec:eval}) as our model selection metric.

\subsection{Evaluation}
We apply five-fold cross validation. For each fold, we run experiments with three seeds. When evaluating performance on the held-out test data for a fold, we use an ensemble of the three models trained with different seeds. Specifically, we take the average prediction across the three models as our final prediction.

\section{Statistical Significance Tests}\label{app:significant-tests}
We conduct statistical significance tests to assess whether the observed differences between methods are statistically meaningful and not attributable to sampling variability. Since we used 5-fold cross-validation, we have five observations per metric for each method. This limited sample size constrains our statistical power, particularly when applying multiple hypothesis test corrections across all pairwise comparisons. We therefore focus our  analysis on key comparisons that identify the best-performing methods.

We use the following statistical significance test: for a given metric (e.g., QWK), we compare two methods using a paired one-sided t-test (paired across test folds to ensure consistent comparisons). We use $\alpha = 0.05$ as the significance threshold. We focus on two hypotheses that help identify the overall best method:
\begin{enumerate}
    \item \textbf{Does CORN outperform OR-Soft in terms of predictive performance?} We find that the difference between the two methods is not statistically significant in terms of QWK $(p = 0.256)$ or MAE $(p = 0.135)$.
    \item \textbf{Does OR-Soft outperform CORN in terms of calibration (ECE)?} We find that OR-Soft achieves significantly lower ECE ($p = 0.025$).
\end{enumerate}

These tests support our main finding that OR-Soft provides the best balance of predictive performance and calibration among the methods we evaluated.

\section{Additional Results}\label{app:additional-results}
\begin{table*}[!t]
\centering
\small
\begin{tabular}{lcc}
\toprule
Method & Cross Entropy & Brier Score \\
\midrule
CE              & 0.93 $\pm$ 0.17 & 0.30 $\pm$ 0.05 \\
CE-Soft         & \underline{0.69 $\pm$ 0.11} & \underline{0.24 $\pm$ 0.05} \\
CORN            & 0.81 $\pm$ 0.16 & 0.25 $\pm$ 0.05 \\
SORD-AE         & 0.95 $\pm$ 0.02 & 0.37 $\pm$ 0.03 \\
SORD-SE         & 0.86 $\pm$ 0.03 & 0.33 $\pm$ 0.02 \\
CORAL           & 1.65 $\pm$ 0.04 & 0.66 $\pm$ 0.03 \\
CORAL-Soft      & 1.62 $\pm$ 0.05 & 0.65 $\pm$ 0.02 \\
OR-CNN          & 0.80 $\pm$ 0.16 & 0.26 $\pm$ 0.04 \\
OR-Soft         & \textbf{0.64 $\pm$ 0.11} & \textbf{0.21 $\pm$ 0.04} \\
\bottomrule
\end{tabular}
\caption{Performance (mean $\pm$ standard deviation) across five folds. Lower is better for both metrics. The best average performance is in \textbf{bold}, second-best is \underline{underlined}. OR-Soft performs best for both metrics, followed by CE-Soft.}
\label{tab:distribution_metrics}
\end{table*}

\begin{table*}[!t]
\centering
\small
\begin{tabular}{lccc}
\toprule
Method & Coverage Error & AUC & Spearman $\rho$ \\
\midrule
CE              & 1.33 $\pm$ 0.08 & 0.90 $\pm$ 0.02 & 0.81 $\pm$ 0.08 \\
CE-Soft         & 1.35 $\pm$ 0.09 & 0.91 $\pm$ 0.04 & 0.77 $\pm$ 0.06 \\
CORN            & \textbf{1.28 $\pm$ 0.10} & \underline{0.92 $\pm$ 0.04} & \textbf{0.84 $\pm$ 0.07} \\
SORD-AE         & 1.34 $\pm$ 0.06 & 0.91 $\pm$ 0.02 & 0.80 $\pm$ 0.06 \\
SORD-SE         & 1.35 $\pm$ 0.09 & 0.90 $\pm$ 0.02 & 0.80 $\pm$ 0.06 \\
CORAL           & 2.54 $\pm$ 0.08 & 0.83 $\pm$ 0.03 & 0.73 $\pm$ 0.09 \\
CORAL-Soft      & 2.51 $\pm$ 0.06 & 0.86 $\pm$ 0.02 & 0.75 $\pm$ 0.06 \\
OR-CNN          & 1.33 $\pm$ 0.10 & 0.92 $\pm$ 0.02 & 0.82 $\pm$ 0.05 \\
OR-Soft         & \underline{1.31 $\pm$ 0.12} & \textbf{0.92 $\pm$ 0.04} & \underline{0.83 $\pm$ 0.05} \\
\bottomrule
\end{tabular}
\caption{Performance (mean $\pm$ standard deviation) across five folds. Lower is better for Coverage Error, higher is better for AUC and Spearman correlation. The best average performance is in \textbf{bold}, second-best is \underline{underlined}. CORN and OR-Soft are the top-performing methods.}
\label{tab:coverage_auc_spearman}
\end{table*}

\begin{figure*}
    \centering
    \includegraphics[width=1\linewidth]{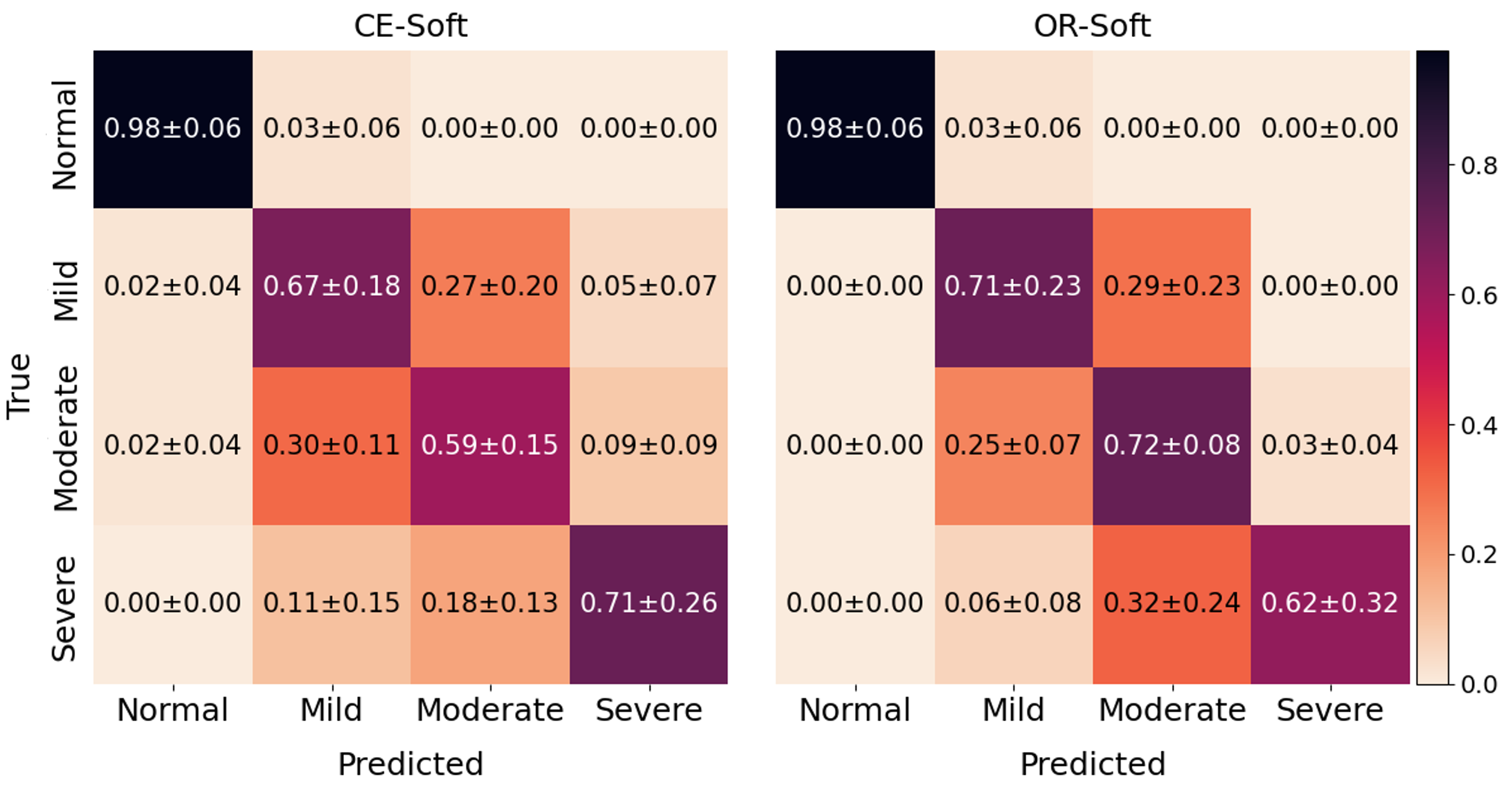}
    \caption{Confusion matrices for CE-Soft and OR-Soft. Confusion matrices are row-normalized. We show mean $\pm$ standard deviation across the five folds. Both methods have high accuracy in discriminating between normal and non-normal cases. OR-Soft makes fewer off-by-two errors than CE-Soft.}
    \label{fig:cm-with-std}
\end{figure*}
\begin{figure*}[htbp]
\floatconts
  {fig:risk-coverage-folds}
  {\caption{Risk–coverage curves for the three methods with the lowest AURC (OR-Soft, OR-CNN, and CORN), shown per test fold.}}
  {%
    \subfigure[Fold 1]{\label{fig:rc-f1}%
      \includegraphics[width=0.32\linewidth]{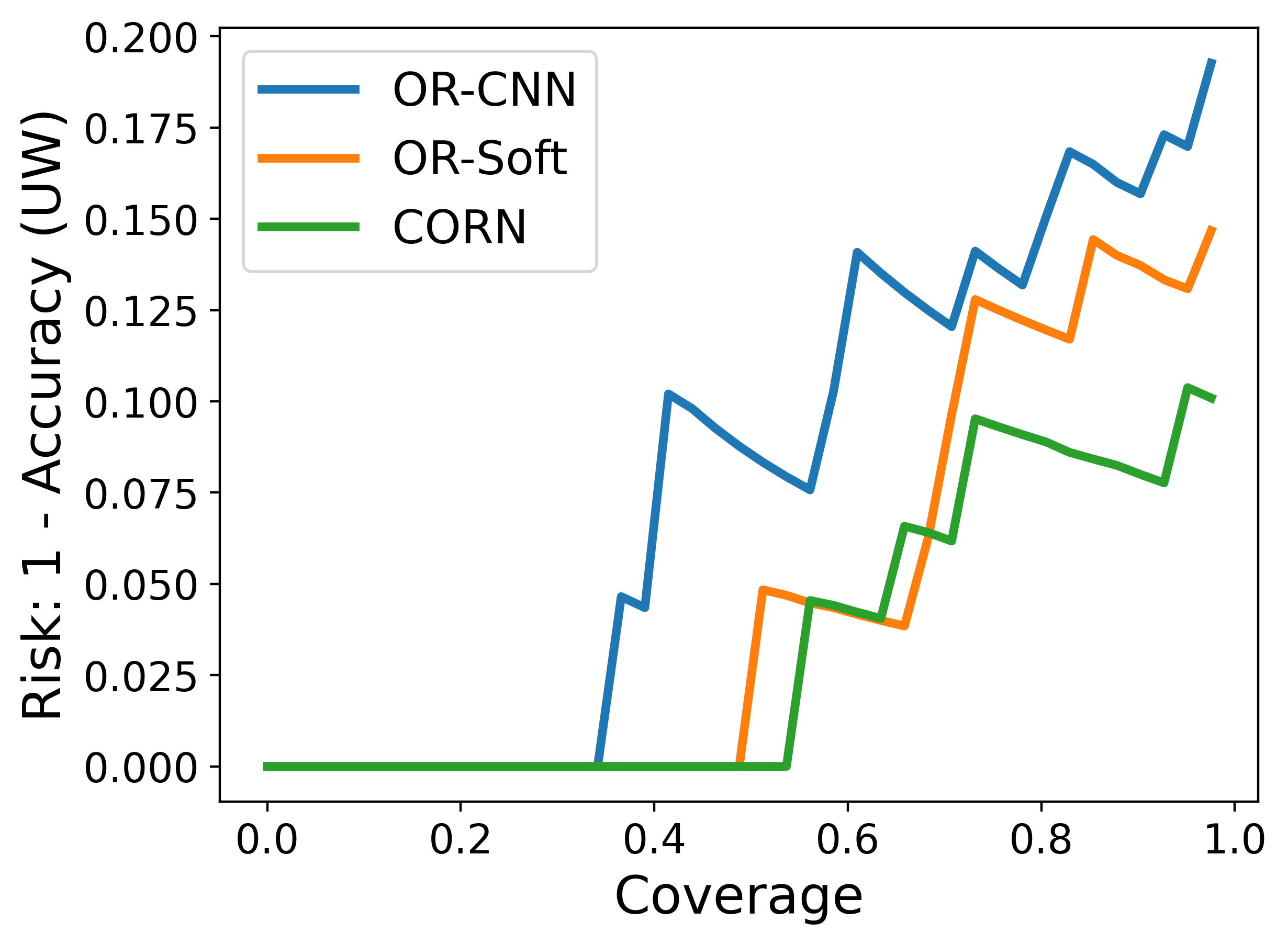}}%
    \hfill
    \subfigure[Fold 2]{\label{fig:rc-f2}%
      \includegraphics[width=0.32\linewidth]{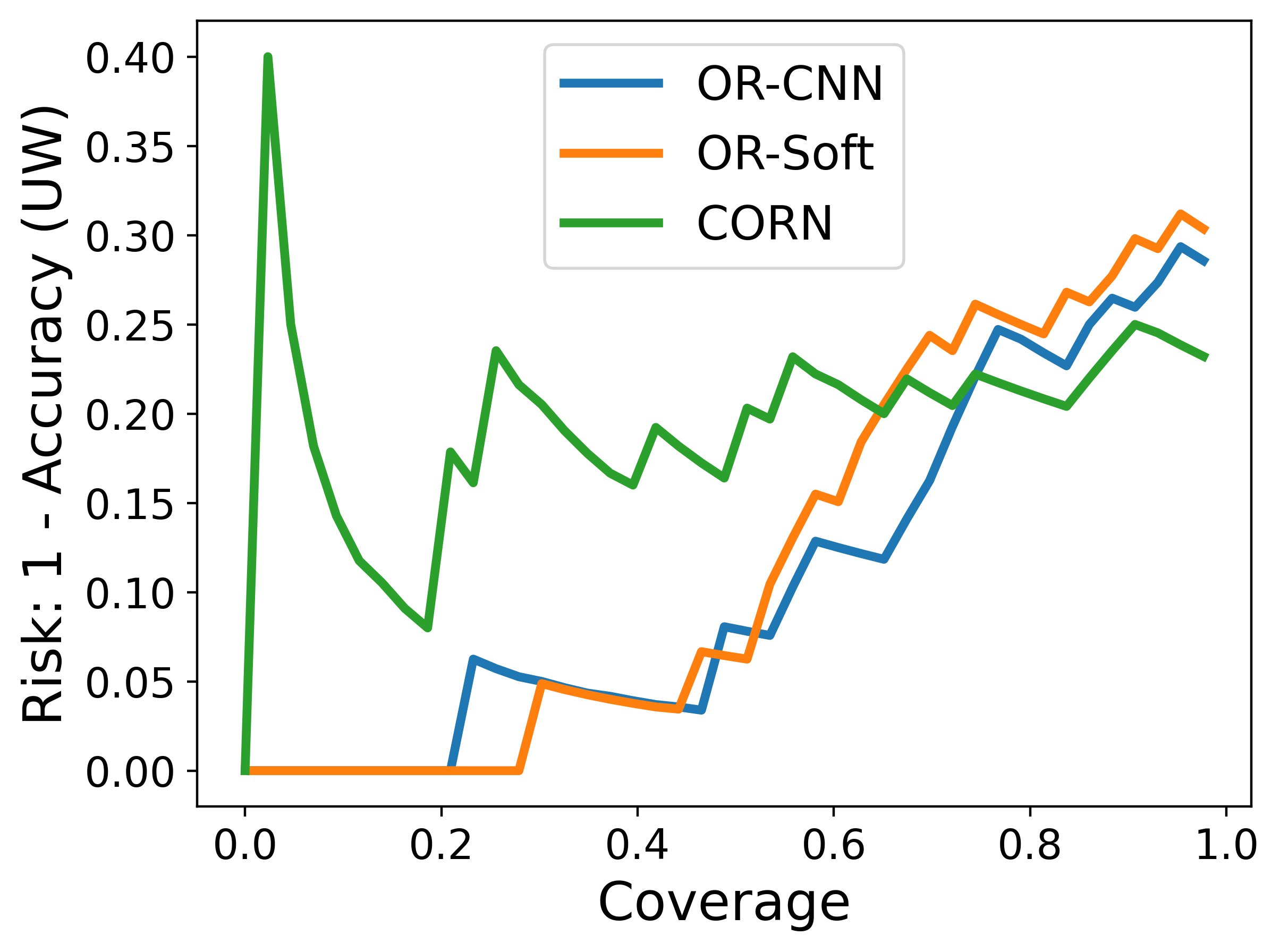}}%
    \hfill
    \subfigure[Fold 3]{\label{fig:rc-f3}%
      \includegraphics[width=0.32\linewidth]{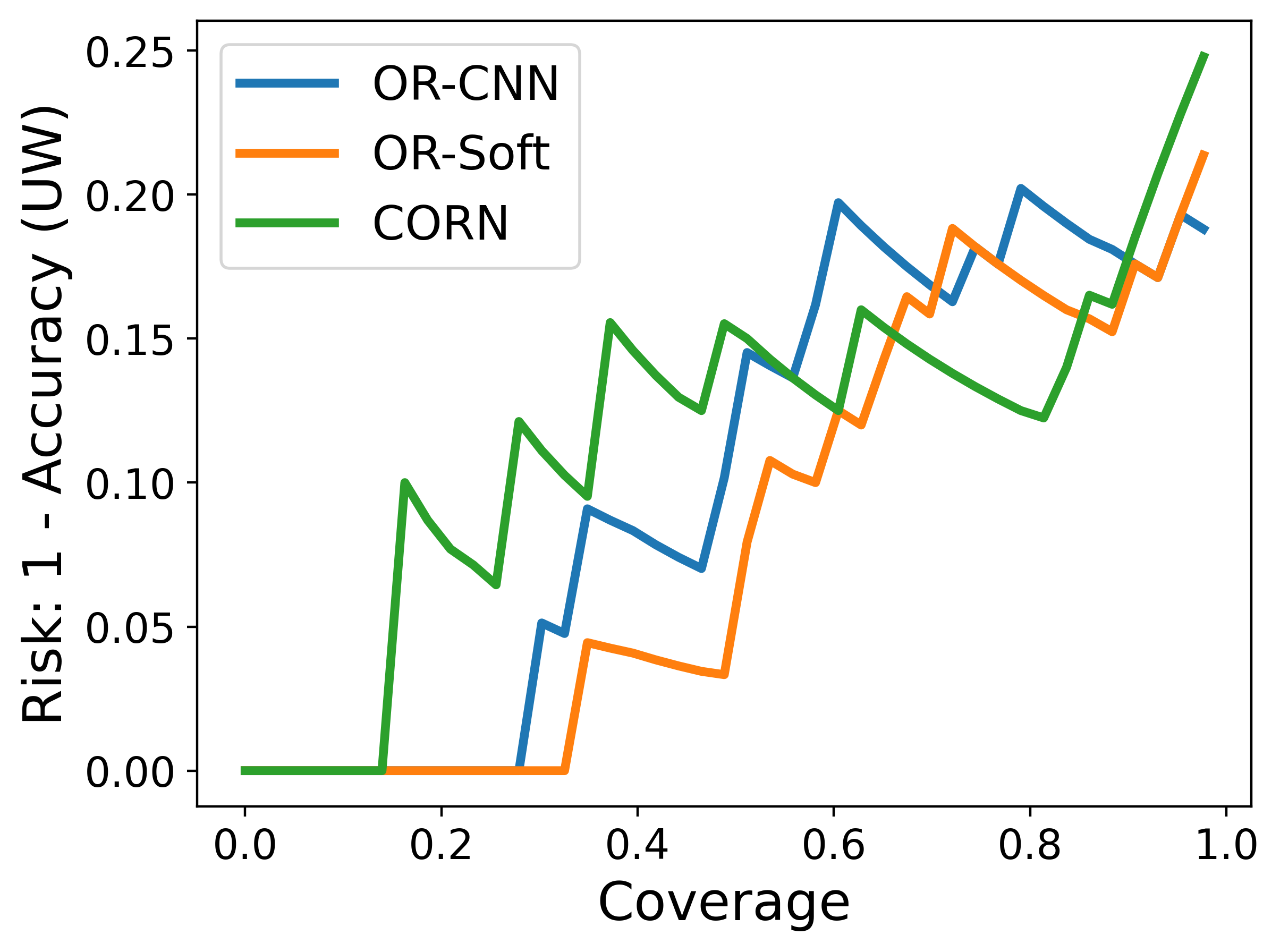}}%

    \par\medskip

    \subfigure[Fold 4]{\label{fig:rc-f4}%
      \includegraphics[width=0.32\linewidth]{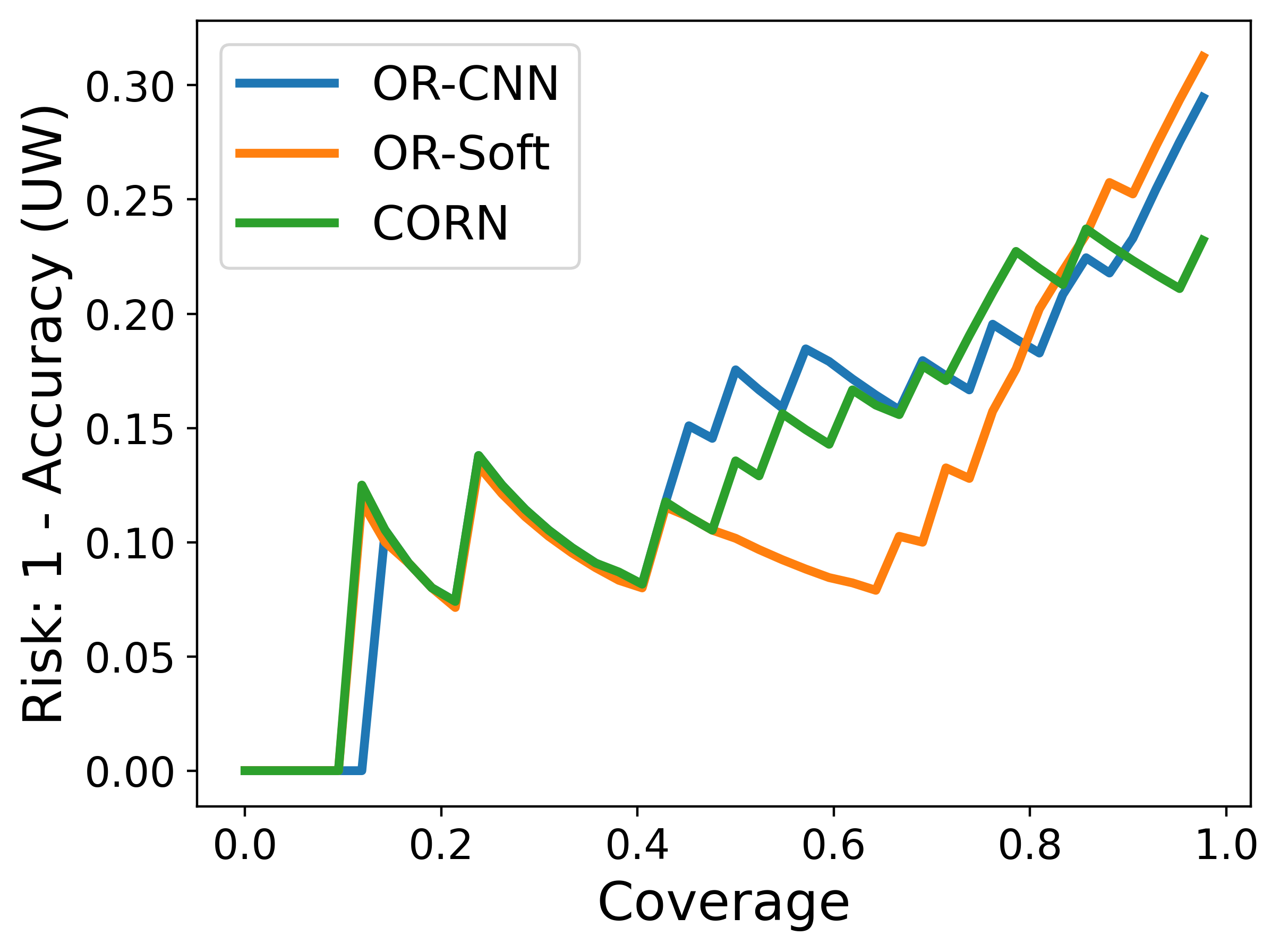}}%
    \subfigure[Fold 5]{\label{fig:rc-f5}%
      \includegraphics[width=0.32\linewidth]{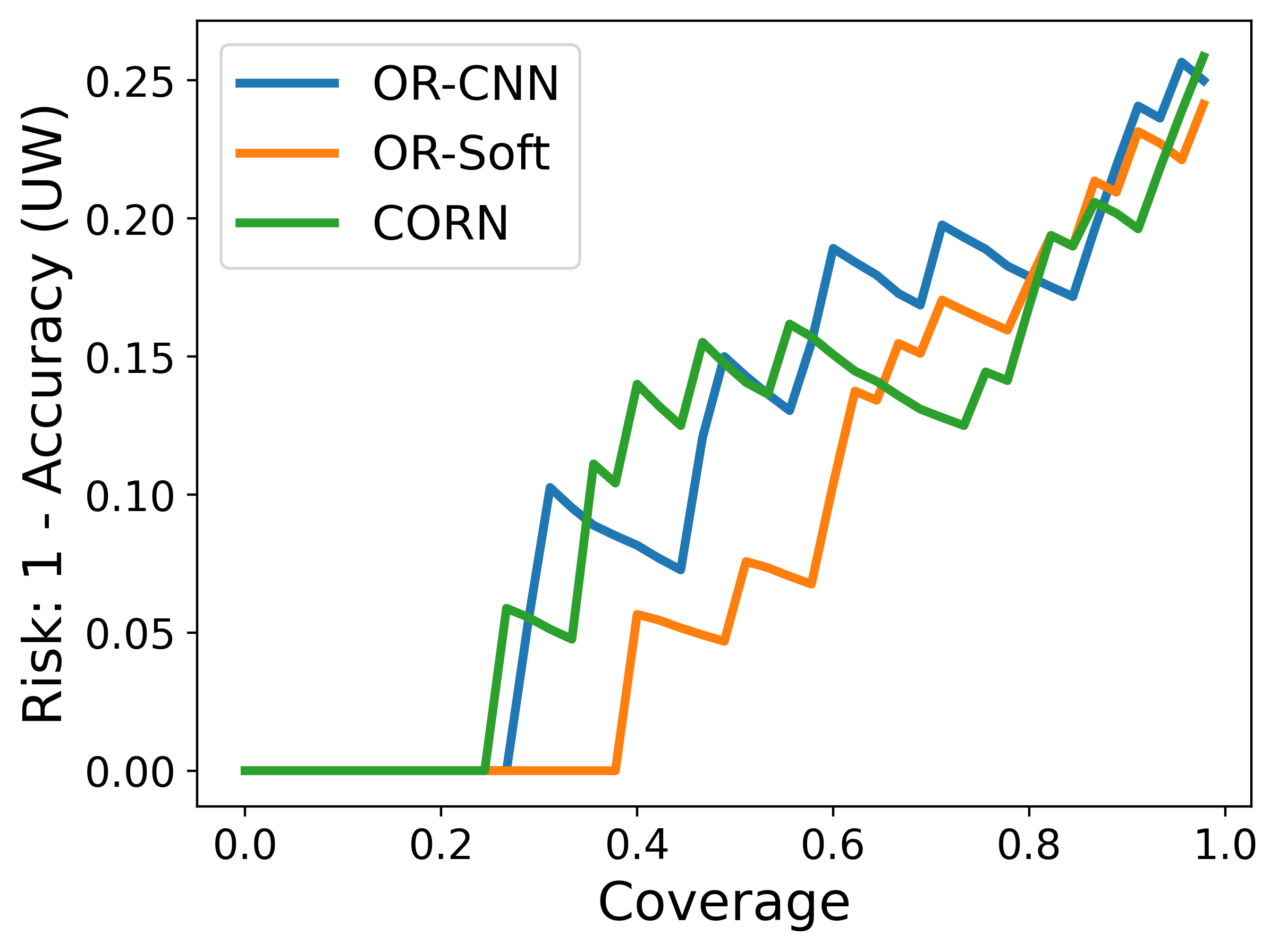}}%
  }
\end{figure*}
\label{fig:risk-coverage-all-folds}

In Table~\ref{tab:distribution_metrics}, we present results for cross entropy and Brier score, two metrics that measure the difference between true and predicted label distributions. In Table~\ref{tab:coverage_auc_spearman}, we present additional metrics that capture predictive performance. We present the same confusion matrices shown in the main text but with the standard deviation across folds included in Figure~\ref{fig:cm-with-std}. We present risk coverage curves for OR-Soft, OR-CNN, and CORN for each individual test fold in Figure~\ref{fig:risk-coverage-folds}.

\section{Predictive Performance Versus Uncertainty Estimation Tradeoff}\label{app:tradeoff}
\begin{figure}
    \centering
    \includegraphics[width=1\linewidth]{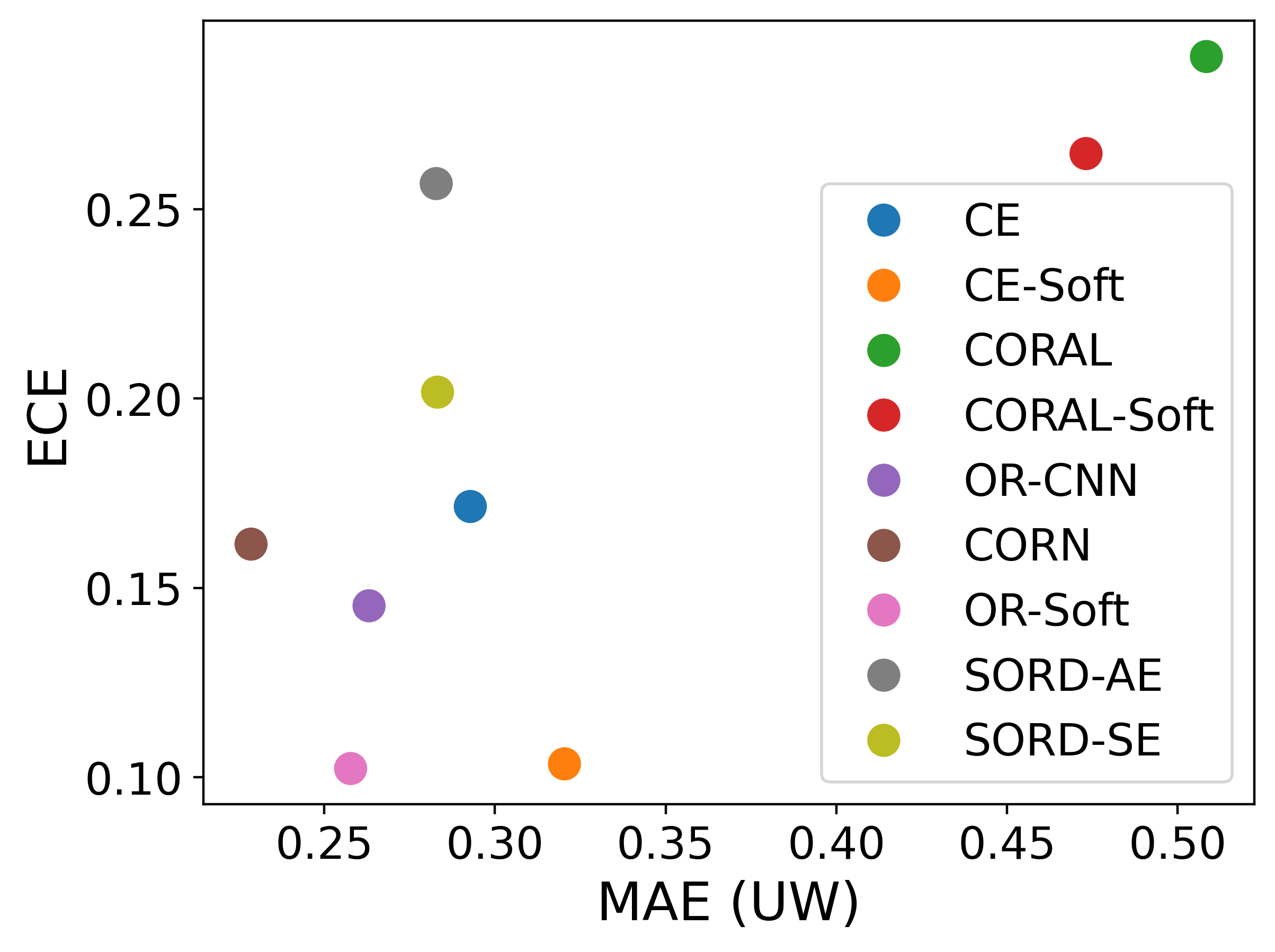}
    \caption{For each method, we plot its predictive performance, measured by MAE (UW), against its uncertainty calibration, measured by ECE. OR-Soft is the in the lower left, with the lowest ECE and a slightly larger MAE (UW) than CORN.}
    \label{fig:tradeoff}
\end{figure}
Among the methods examined in this study, we observed a tradeoff between predictive performance and uncertainty estimation quality. To better understand this tradeoff, in Figure~\ref{fig:tradeoff}, we present a scatter plot with MAE (UW) on x-axis and ECE on the y-axis showing where each method falls along this tradeoff.

\section{Results from Full Multi-Rater Dataset}\label{app:full-data}
\begin{table*}[!t]
\centering
\small
\begin{tabular}{lcccccc}
\toprule
Method & MAE (UW) & QWK (UW) & Accuracy (UW) & Accuracy (AR) & ECE & AURC \\
\midrule
CE      & 0.33 $\pm$ 0.13 & 0.76 $\pm$ 0.12 & 0.69 $\pm$ 0.11 & 0.88 $\pm$ 0.06 & 0.21 $\pm$ 0.04 & 0.18 $\pm$ 0.05 \\
CE-Soft & 0.34 $\pm$ 0.10 & 0.77 $\pm$ 0.07 & 0.67 $\pm$ 0.10 & 0.86 $\pm$ 0.05 & \underline{0.15 $\pm$ 0.02} & 0.22 $\pm$ 0.07 \\
CORN    & 0.30 $\pm$ 0.08 & 0.80 $\pm$ 0.09 & 0.72 $\pm$ 0.07 & \underline{0.89 $\pm$ 0.05} & 0.17 $\pm$ 0.04 & 0.13 $\pm$ 0.05 \\
OR-CNN  & \underline{0.30 $\pm$ 0.07} & \underline{0.81 $\pm$ 0.08} & \textbf{0.73 $\pm$ 0.04} & \textbf{0.91 $\pm$ 0.02} & 0.18 $\pm$ 0.04 & \underline{0.13 $\pm$ 0.03} \\
OR-Soft & \textbf{0.28 $\pm$ 0.06} & \textbf{0.83 $\pm$ 0.03} & \underline{0.73 $\pm$ 0.06} & 0.88 $\pm$ 0.05 & \textbf{0.10 $\pm$ 0.01} & \textbf{0.13 $\pm$ 0.05} \\
\bottomrule
\end{tabular}
\caption{Results on the dataset with annotations from three additional clinical experts (obtained after the submission deadline). Performance (mean $\pm$ standard deviation) across five folds is shown. The method with the best average performance is in \textbf{bold}, second-best is \underline{underlined}. UW = Uncertainty-Weighted, AR = Any-Rater Accuracy (see Section~\ref{subsec:eval}). OR-Soft performs best in terms of both ordinal metrics (MAE, QWK) and uncertainty estimation metrics (ECE, AURC).}
\label{tab:methods_soft_hard_6_rater}
\end{table*}

After the submission deadline, we collected annotations from three additional laryngeal surgeons for all subjects in the dataset. We conducted a preliminary analysis on the data with this new label set; results are shown in Table~\ref{tab:methods_soft_hard_6_rater}. 

In this analysis, we took the annotations provided by the three new raters and combined them with the labels used in our initial analysis. For images that had annotations from three raters in the initial set (i.e, the Multi-Rater subset), we directly combined these with the three new annotations, yielding six ratings per image. For images in the initial set that did not have three independent ratings -- specifically, the normal cases identified through comprehensive clinical screening and the severe cases labeled by a three-person consensus -- we replicated their initial labels three times to maintain balanced weighting across the two annotation sets. This approach ensured that all images had exactly six ratings.

We derived soft labels from the empirical distribution over these six ratings. We created hard labels by choosing the mode rating. For some images, there was a 50-50 tie between two classes. We excluded these images from evaluation but retained them for training. To train with these images, for soft-label approaches, we used the full six-rater distribution directly. For hard label approaches, we randomly sampled from the majority classes at each training epoch.

The results in Table~\ref{tab:methods_soft_hard_6_rater} largely align with those presented in the main text. OR-Soft achieves the best balance between predictive performance and uncertainty estimation, and the soft label methods have superior uncertainty calibration compared to their hard label variants. One difference is that OR-Soft performs slightly better than CORN in terms MAE, QWK, and Accuracy in this analysis. However, the results are not statistically significant ($p\geq 0.21$ for each of these metrics), which is consistent with our main findings.

\end{document}